\documentclass{article}
\usepackage{spconf,amsmath,graphicx}
\usepackage{algorithm}
\usepackage[noend]{algpseudocode}% Example definitions.
\usepackage{hyperref}
\usepackage{multirow}
\usepackage{balance}
\title{BEAM SEARCH DECODING USING MANNER OF ARTICULATION DETECTION KNOWLEDGE DERIVED FROM CONNECTIONIST TEMPORAL CLASSIFICATION}
\name{Pradeep R and Sreenivasa Rao K\thanks{We thank Tata Consultancy Services (TCS) for sponsoring the research under TCS-Research Scholar Programme.}}
\address{Department of Computer Science and Engineering,\\ Indian Institute of Technology, Kharagpur, India \\ {\small \tt \{pradeep\_raj31, ksrao\}@iitkgp.ac.in}
}
\begin{document}
%\ninept
%
\maketitle
\begin{abstract}
Manner of articulation detection using deep neural networks require a priori knowledge of the attribute discriminative features or the decent phoneme alignments. However generating an appropriate phoneme alignment is complex and its performance depends on the choice of optimal number of senones, Gaussians, etc. In the first part of our work, we exploit the manner of articulation detection using connectionist temporal classification (CTC) which doesn't need any phoneme alignment. Later we modify the state-of-the-art character based posteriors generated by CTC using the manner of articulation CTC detector. Beam search decoding is performed on the modified posteriors and it's impact on open source datasets such as AN4 and LibriSpeech is observed.
\end{abstract}
\begin{keywords}
Manner of articulation, connectionist temporal classification, speech recognition
\end{keywords}
\section{Introduction}
\label{sec:intro}

Deep neural networks (DNNs) combined with hidden Markov models (HMM) have become the dominant approach in acoustic modeling \cite{dahl2012context} and substantial error rate reduction has been achieved for speech recognition tasks \cite{maas2017building}. Recurrent neural networks (RNNs) have performed better than traditional DNNs, because they can detect events outside of a fixed temporal window size and are less affected by temporal distortion. For these reasons, Long Short-Term Memory RNNs (LSTM RNNs) \cite{zhang2016highway} are more suitable for sequence tasks such as sequence modeling , and have been helpful to improve robustness in speech recognition \cite{greff2017lstm} .

Substantial focus on ASR happened by the introduction of a a simple paradigm to perform speech recognition in an end-to-end manner. The two major end-to-end ASR implementations based on both connectionist temporal classification (CTC) \cite{graves2006connectionist} \cite{graves2014towards} and attention-based encoder-decoder network \cite{bahdanau2016end}. CTC uses Markov assumptions to efficiently solve sequential problems by dynamic programming.  On the other hand, the attention-based method \cite{graves2006connectionist} exploits an attention mechanism to perform alignment between acoustic features and recognized symbols. However, the basic temporal attention mechanism is too flexible in the sense that it allows extremely non-sequential alignments.
This is rational for applications such as machine translation where input and output word order are different \cite{bahdanau2014neural}. However in phone attribute detection, the acoustic features and the corresponding outputs proceed in a monotonic way. Since CTC permits an efficient computation of a strictly monotonic alignment using dynamic programming, we propose to train a CTC-based manner of articulation detector to detect vowels, semi-vowels, nasals, fricatives and stop consonants.

Speech attributes were detected using discriminative features at the front-end and training the classifier \cite{lee2007overview}. Signal processing approaches are used for automatic and accurate detection of the closure-burst transition events of stops and affricates \cite{ananthapadmanabha2014detection}. Later deep learning techniques were used to detect speech attributes \cite{siniscalchi2013exploiting}.
However, training such complex systems involves feature extraction, phoneme force alignment and deep neural network training. Recently, Cernak et al., \cite{cernak2018nasal} exploited a solution to train a nasal sound detector without phone alignment using an end-to-end phone attribute modeling based on the connectionist temporal classification.

Recent success on nasal detection using CTC motivated us to extend their framework for detecting five broad manners of articulation namely vowel, semi-vowel, nasal, fricative and stop consonant. In the first part of our work, we extend CTC based nasal and non-nasal detection framework \cite{cernak2018nasal} to detect five broad manners of articulation. Later we modify the state-of-the-art character based posteriors generated by CTC using the manner of articulation CTC detector. Beam search decoding \cite{freitag2017beam}  is performed on the modified posteriors and it's impact on open source datasets such as AN4 and LibriSpeech is observed.

% The rest of the paper is organized as follows: Section \ref{sec:mannerDetect} overviews the proposed CTC based manner of articulation detector, Section \ref{sec:charCTCfromManner} illustrates the method of obtaining modified character CTC using manner CTC, Section \ref{sec:Experiments} introduces the experimental setup used for evaluation, and Section \ref{sec:Results} presents results on various alternatives of the proposed detector described in Section \ref{sec:charCTCfromManner}. Section \ref{sec:conclude} summarises and concludes the paper.

\section{Manner of Articulation CTC detector}
\label{sec:CTCMannerDetect}

For a given input $X$, CTC gives us an output distribution over all possible $Y$'s. We can use this distribution either to infer a likely output or to assess the probability of a given output. CTC allows repetitions of output labels
and extends the set of target labels with an additional blank
symbol, which represents the probability of not emitting any
label at a particular time step. It introduces an intermediate
representation called the CTC path. A CTC path is a sequence
of labels at the frame level, allowing repetitions and the blank
to be inserted between labels.

To be precise, the CTC objective for a single $(X,Y)$ pair is:
\begin{eqnarray}
p(Y|X) = \sum_{A\in A_{X,Y}}\prod_{t=1}^{T}p_t(a_t|X)
\end{eqnarray}
The conditional probability marginalizes over the set of valid alignments by computing the probability for a single alignment step-by-step. The conditional probability of the labels at each time step, $p_t(a_t|X)$, is generally estimated using a RNN (LSTM/GRU).

%Figure \ref{fig:OverallDecoding} 
\begin{figure}[htb]
\begin{minipage}[b]{1.0\linewidth}
  \centering
  \centerline{\includegraphics[width=8.5cm,height=4.0cm]{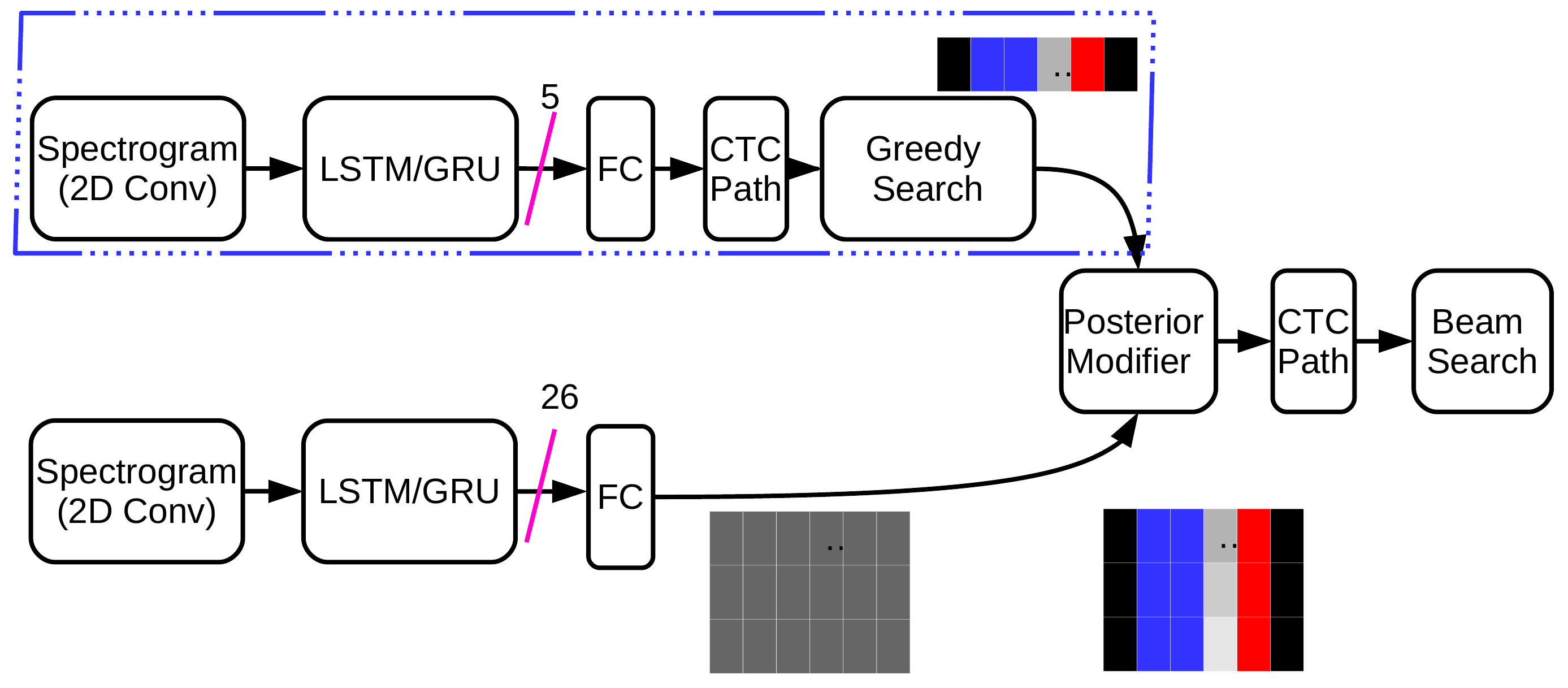}}
%  \vspace{2.0cm}
  %\centerline{(a) Result 1}\medskip
  \caption{Character CTC derived from manner of articulation CTC}
  \label{fig:proposedOverview}
  %\vspace{-0.8cm}
\end{minipage}
\end{figure}

In order to detect manners of articulation using CTC, the softmax output nodes are set to $M+k$ where $M$ is is the number of manners and $k$ characters are added to include blank ($<$) or space ($>$). The character level transcripts are mapped to five different manners of articulation \{$V, \$, N, F, S$\} that represents vowel, semi-vowel, nasal, fricative and stop consonant respectively. The manner of articulation detector using CTC is shown in upper part of Figure \ref{fig:proposedOverview}. The left part of the network is two layers of convolutions over both time and frequency domains. Temporal convolution is commonly used in speech processing to efficiently model temporal invariance for variable length utterances. Convolution in frequency attempts to model spectral variance due to speaker variability and it has been shown to further improve the performance \cite{sainath2013deep}. Following the convolutional layers are bidirectional recurrent layers.
After the bidirectional recurrent layers, a fully connected (FC) layer is applied and the output is produced through a softmax function computing a probability distribution over the target labels {blank, vowel, semi-vowel, nasal, fricative, stops, space}. The model is trained using the CTC loss function. To accelerate the training procedure, Batch Normalization \cite{ioffe2015batch} is applied on hidden layers.

\section{Beam Search Decoding on Modified Posteriors}
\label{sec:pagestyle}

The proposed method is illustrated in Figure \ref{fig:proposedOverview}. The manner of articulation portions are detection using CTC using the procedure described in Section \ref{sec:CTCMannerDetect}. To create a manner transcription, we use the training procedure of \cite{graves2014towards} with appropriate nodes at the softmax layer and greedily search the best path $p \in L^T$ :
\begin{eqnarray}
\underset{p}{\arg\max}\ \prod_{t=1}^{T} P_{AM}^t(p_t|X)
\end{eqnarray}
where  $L$ is the augmented label set for each frame.
The mapping of the path to a transcription $z$ is straight forward and works by applying the squash function: $z = B(p)$. For manner based CTC acoustic models this procedure can provide CTC peaks that correspond to manner of articulation segments in speech. The output of the manner CTC detector is the most probable manner index as illustrated in different colors in top part of Figure \ref{fig:proposedOverview}. We describe the black , blue and red portions as a blank symbol, vowel and semi-vowel for illustration. On the other hand, the character CTC as described in lower part of Figure \ref{fig:proposedOverview} generates the posteriors of size say (26+k) $\times$ $T$ where $T$, $k$ are the number of frames in the test utterance and the additional characters such as space, ', etc., used in training CTC.

\subsection{Posterior Modification using Manner CTC Detector}

For each frame in the character posterior, we observe the corresponding most probable manner of articulation index generated using Best path/Greedy search. The manner index is initially mapped to all possible character index as described in Table \ref{tab:manner2charTable}. For instance the vowel manner ($V$) can be from any of the five characters say $A, E, I, O, U$ and hence it is mapped to a cell that contains corresponding character index say $1, 5, 9, 15, 21$ respectively as given in Table \ref{tab:manner2charTable}.
\begin{table}[h]
\centering
\caption{\it Manner of Articulation to Character Index}
\footnotesize\setlength{\tabcolsep}{1.8pt}
\renewcommand{\arraystretch}{1.0}
\label{tab:manner2charTable}
\begin{tabular}{|l|l|l|}
\hline
\textbf{Manner CTC Targets} & \textbf{Character CTC Targets} & \textbf{Character Index} \\ \hline
- & - & 0 \\ \hline
V & A, E, I, O, U & 1, 5, 9, 15, 21 \\
\$ & L, R, W, Y & 12, 18, 23, 25 \\
N & M, N & 13, 14 \\
F & F, H, J, S, V, X, Z & 6, 8, 10, 19, 22, 24, 26 \\
S & B, C, D, G, K, P, Q, T & 2, 3, 4, 7, 11, 16, 17, 20 \\ \hline
\textgreater{} & \textgreater{} & 27 \\ \hline
\end{tabular}
\end{table}

For a frame that belongs to a vowel manner, the posteriors of the vowel character indices are retained whereas other non-vowel posteriors for that frame are forced to zero. This will ensure that the maximum posterior probability of vowel manner is forced to generate only the vowel characters. The modified posterior generated is illustrated in lower part of Figure \ref{fig:proposedOverview}. The different colors in the modified posteriors imply that only the character indices corresponding to the most probable manner index are active at frame $t$.

\subsection{Combining Manner CTC and Character CTC Detectors}

The basic version of obtaining manner based character CTC is shown in Algorithm 1.
\makeatletter
\def\BState{\State\hskip-\ALG@thistlm}
\makeatother

\begin{algorithm}
\caption{: $Result \gets MannerbasedCharCTC(Data)$}\label{euclid}
\begin{algorithmic}[1]
\State ${\textbf{Data}: postManner, postChar, labelsChar}$
\State ${\textbf{Result}: decoString}$
\State $mannerInx \gets argmax(postManner)$
\State $mcInx \gets manner2char(mannerInx)$
\For {$frame = 1 : length(mannerInx)$} :
\State $nonMannerInx \gets find(labelsChar \neq mcInx({frame})$
\State $postC(frame,nonMannerInx) \gets 0 $
\State $newPostC(:,frame) \gets normalize(postC(:,frame))$
\EndFor
\State $decoString \gets BeamSearch(newPostC,labelsChar)$
\State $\textbf{return}$  $decoString$
%\BState \emph{loop}:
%\If {$\textit{string}(i) = \textit{path}(j)$}
%\State $j \gets j-1$.
%\State $i \gets i-1$.
%\State \textbf{goto} \emph{loop}.
%\State \textbf{close};
%\EndIf
%\State $i \gets i+\max(\textit{delta}_1(\textit{string}(i)),\textit{delta}_2(j))$.
%\State \textbf{goto} \emph{top}.
%\EndProcedure
 % \label{fig:mannerbasedCTC}
\end{algorithmic}
\end{algorithm}
 We find the frame level index of the most probable manner of articulation portions obtained from posteriors manner (postManner) (line 3). The manner to character indices (mcInx) are generated according to Table \ref{tab:manner2charTable} (4).  We iteratively find the index of the most probable character segment based on manner CTC detector knowledge. For every frame, we find the non manner character indices and force the posterior probabilities to zero (6, 7). The character posteriors in the appropriate manner portions are normalized to form modified character posteriors, newPostC (8). Finally the conventional beam search technique \cite{freitag2017beam} is applied on the modified posteriors to decode the most optimal sequence (9). The beam search mechanism chooses beam of $B$ hypotheses at every frame and iteratively modifies the successive posterior probabilities depending on the blank or non-blank probabilities at that time instant.

The basic idea of the proposed method is to force the CTC to generate appropriate character label according to the manner of articulation knowledge. The advantage of using such technique is that symbols which are missed out in the baseline character CTC but present in the manner CTC are forced to emit some symbol at that frame. Hence it may impact in reduction of some of the deletion, substitution or insertion errors as compared to that of state-of-the-art decoding method.

\section{Experiments}
\label{sec:Experiments}

\subsection{Data}

We used two open source databases for training the manner and character CTC systems and are illustrated below:(1) AN4 \footnote{http://www.speech.cs.cmu.edu/databases/an4/} \-- the database contains alpha numeric speech data having 948 training and 130 test utterances. The dataset provides a good sample to achieve deterministic results to scale up with larger datasets.
(2) LibriSpeech \footnote{http://www.openslr.org/resources/12/} \-- the data are sampled at 16 kHz, and the training part of the corpus is split into three sub-sets, with size approximately 100, 360 and 490 hours respectively. In our experiments, we use 100 hours train-clean corpus.

\subsection{Training}

The training phase is based on the open source DeepSpeech2 \footnote{https://github.com/SeanNaren/deepspeech.pytorch} architecture \cite{amodei2016deep}, trained with the CTC activation function.
  The manner detector using CTC starts with two layers of 2D convolutions over both time and frequency domains with 32 channels, 41 $\times$ 11, 21 $\times$ 11 filter dimensions, and 2 $\times$ 2, 2 $\times$ 1 stride. Four next bidirectional gated recurrent layers with 400 hidden units are followed by one fully connected linear layer with 7 softmax outputs \{$blank,', vowel, semi-vowel, nasal, fricative , stop, space$\}. The GRU models have around 4.1 millions (M) parameters. The input sequence are values of spectrogram slices, 20 ms long, computed from Hamming windows with 10 ms frame shifts. The input sequences are thus the values of the natural logarithm of one plus the magnitude components of the short-time Fourier transform of the windowed input signal.
The word transcription of the input signal was used to prepare the output sequences. The output (target) sequence was obtained directly from the letters of the word transcription. 
%As all the nasal sounds in our data [m,n,N] can be produced by pronouncing only the letters M and N, the output sequence was prepared without any lexicon, where all the letters M and N are converted into the nasal labels, all the other letters into the nonasal labels, and their repetitions were replaced by the single label.
 The space denoted the word boundary. We used 50 epochs to train all the models used for further evaluation.

\section{Results}
\label{sec:Results}

We evaluated both character and manner CTC detectors on the test-clean data set. 
We computed then the manner of articulation posterior probabilities by running forward pass of the manner CTC detector. The manner of articulation error  rate (MER) is used to measure the performance of the manner CTC detector. The calculation of MER is similar to that of character error rate where the manner based reference and the obtained transcripts are compared. The reference transcripts is initially changed by converting each character in the transcription to the appropriate manner label. 
Table \ref{tab:MER} shows obtained MERs of the CTC manner of articulation detector obtained on three datasets.
\begin{table}[h]
\centering
\caption{\it Manner of Articulation Error Rate for Manner CTC detector }
\footnotesize\setlength{\tabcolsep}{4pt}
\renewcommand{\arraystretch}{1.0}
\label{tab:MER}
\begin{tabular}{|c|c|}
\hline
\textbf{Dataset} & \textbf{\% MER} \\ \hline
AN4 & 2.8 \\ \hline
LibriSpeech & 2.7 \\ \hline
\end{tabular}
\end{table}

Table \ref{tab:WERCER} shows the  word error rate (WER) and the character error rate (CER) obtained using the baseline and the proposed method.
\begin{table}[h]
\centering
\caption{\it WER and CER obtained using baseline CTC and the proposed method on different datasets}\footnotesize\setlength{\tabcolsep}{6pt}
\renewcommand{\arraystretch}{1.0}
\label{tab:WERCER}
\begin{tabular}{|c|c|c|c|}
\hline
\textbf{Dataset} & \textbf{Method} & \textbf{\% WER} & \textbf{\%CER} \\ \hline
\multirow{2}{*}{AN4} & Baseline & 9.3 & 3.7 \\ \cline{2-4} 
 & Proposed & 8.7 & 3.0 \\ \hline
\multirow{2}{*}{LibriSpeech} & Baseline & 11.1 & 3.3  \\ \cline{2-4} 
 & Proposed & 10.7 & 2.9 \\ \hline
\end{tabular}
\end{table}
The pre-trained manner of articulation models and the baseline CTC models trained with AN4 dataset is made as an open source code\footnote{https://github.com/Pradeep-Rangan/Manner-of-Articulation-Detection-using-CTC}.
It is observed that the manner of articulation knowledge in modifying the CTC path has significant impact in improving the performance of ASR.

\subsection{Discussion}

The CNN in the used model performed 2D convolution, where the first dimension is frequency and the second dimension is time. A longer stride is usually applied to speed-up training. Using the stride in the time dimension results into time compacting of the input audio, e.g., using the stride of 2 results into 2 times less frames of the output. For applications where time alignment is required, we experimented with the stride of 1. The training takes twice longer as with the stride 2, but for this phone-attribute task the training still converges well.

%Figure \ref{fig:OverallDecoding} 
\begin{figure}[htb]
\begin{minipage}[b]{1.0\linewidth}
  \centering
  \centerline{\includegraphics[width=8.0cm,height=6.0cm]{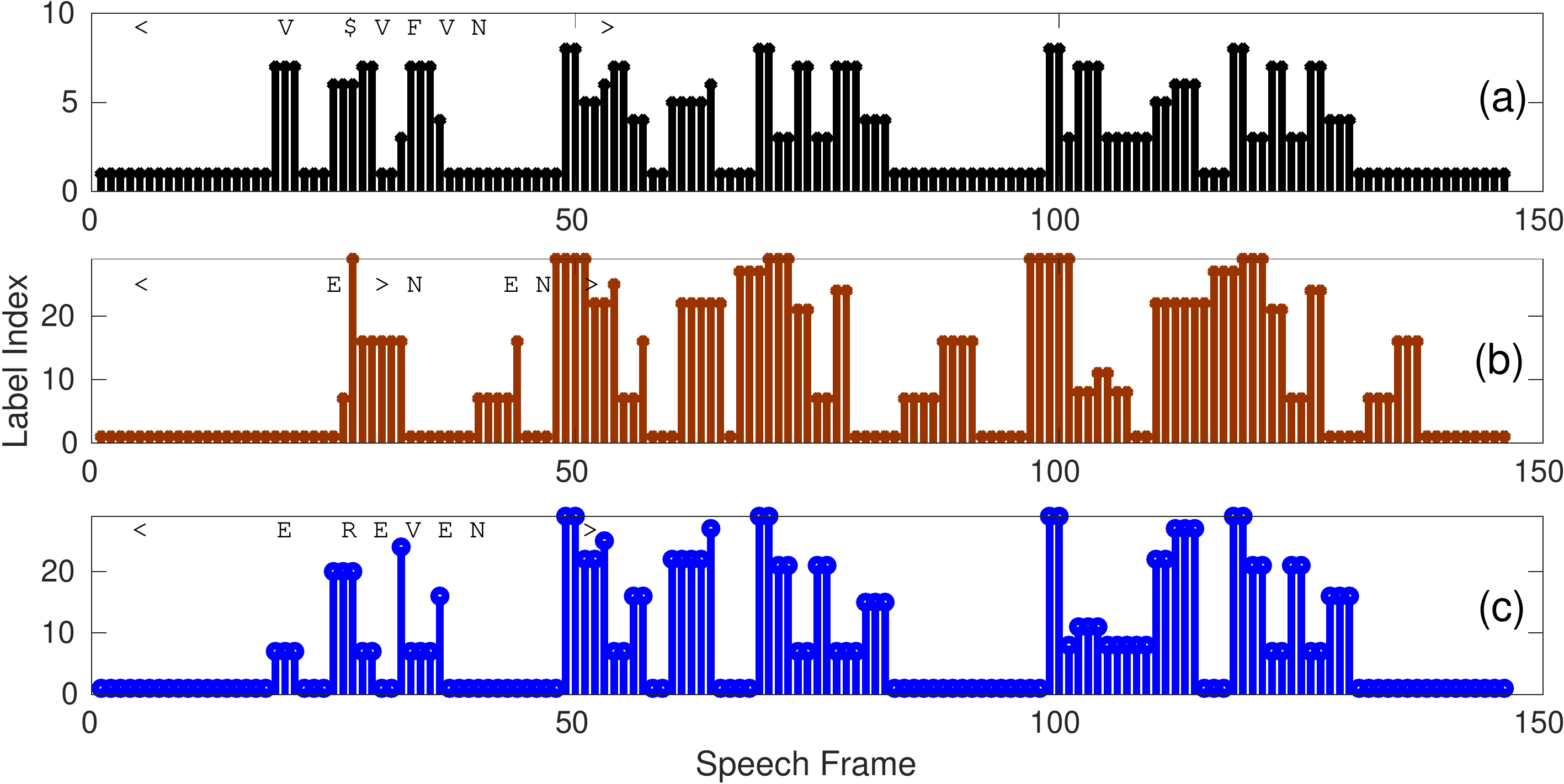}}
%  \vspace{2.0cm}
  %\centerline{(a) Result 1}\medskip
  \caption{Illustration of manner of articulation CTC on character CTC (a) manner of articulation Index (b) baseline character index (c) modified manner based character index}
  \label{fig:compareResults}
 % \vspace{-0.8cm}
\end{minipage}
\end{figure}

Figure \ref{fig:compareResults} shows an example of the label index generated on manner CTC, baseline character CTC and the modified manner based character CTC detector.
The speech utterance is from AN4 test dataset $(an4dataset/test/an4/wav/cen8-fcaw-b.wav)$ whose content has the sentence ``ELEVEN TWENTY SEVEN FIFTY SEVEN''. The most probable manner index is derived from the posteriors manner as shown in Figure \ref{fig:compareResults} (a). On top of the figure we illustrate some of the text transcript portions. The baseline character CTC as shown in Figure \ref{fig:compareResults} (b) generates ``E NEN TWENTY SEVEN FIFTY SEVEN'' leading to false insertion and substitution errors.
The advantage of using such technique is that symbols which are missed out in the baseline character CTC but present in the manner CTC are forced to emit some symbol at that frame. 
  Figure \ref{fig:compareResults} (c) shows the modified character index . The decoded sequence obtained using proposed method is : ``EREVEN TWENTY SEVEN FIFTY SEVEN''. It can be observed that the additional space that was generated using baseline method is nullified using the proposed method. Also the blank character propbabilties that dominated to miss out the substring `EVEN' is recovered. The generation of the character `R' may be due to the fact that the manner of articulation has semivowel. The probabilty of occurrence of character `L' is less than that of `R'.

%DNN and LSTM, us-
%ing stride 1 in the time domain, nasal detectors for a sample
%librivox recording. Firstly, the CTC output is peaky, which
%suggests that the CTC method can better deal with nasality
%characteristics due to coarticularion and better separate them
%from nasality associated to the features of the phoneme, than
%the DNN based method. The peaks slightly change the po-
%sition with individual training epochs. Secondly, the CTC
%output contains less false positives that propagates in over-
%all better performance, comparing to the DNN detector, in the
%terms of EERs. We speculate that these CTC nasal detector
%properties cause the lower detection thresholds observed dur-

%\vfill\pagebreak

\section{Conclusion}
\label{sec:conclude}

This paper has proposed to use the connectionist temporal classification for the end-to-end manner of articulation modeling.
The manner of articulation knowledge is deployed in the conventional character CTC path to regenerate the new character CTC path. The modified manner based character CTC is evaluated on open source speech datasets such as AN4 and LibriSpeech and it outperforms over the baseline character CTC.
Application of the proposed manner of articulation CTC detector in weight adaptation of baseline end-to-end ASR training is also planned for future work.

\bibliographystyle{IEEEbib}
\balance
\bibliography{strings,refs}

\end{document}